\documentclass[letterpaper, 10 pt, conference]{ieeeconf}  
\IEEEoverridecommandlockouts                              
\usepackage{url}
\usepackage{amsmath,amssymb}
\usepackage{graphicx,color}
\usepackage{booktabs}
\usepackage{multirow}
\usepackage{verbatim}
\usepackage{gensymb}
\usepackage{pythonhighlight} 
\graphicspath{{figures/}}
\newcommand{\ba}{\mathbf{a}}

\newcommand{\bo}{\mathbf{o}}

\newcommand{\ie}{i.e., }
\newcommand{\eg}{e.g., }

\usepackage{color, soul}
\renewcommand\hl[1]{#1} 

\usepackage{siunitx} 
\DeclareMathOperator*{\argmax}{arg\,max}

\usepackage{pifont}
\newcommand{\cmark}{\textcolor[HTML]{59a14f}{\ding{51}}}%
\newcommand{\xmark}{\textcolor[HTML]{e15759}{\ding{55}}}%

\definecolor{britishracinggreen}{rgb}{0.23, 0.33, 0.19}

\usepackage[font={footnotesize}]{caption}
\def\tablename{Table}

\title{\LARGE \bf
Learning to Rearrange Deformable Cables, Fabrics, and Bags\\with Goal-Conditioned Transporter Networks
}

\author{Daniel Seita$^{1,*}$, Pete Florence$^{2}$, Jonathan Tompson$^2$, \\ Erwin Coumans$^2$, Vikas Sindhwani$^2$, Ken Goldberg$^1$, Andy Zeng$^{2}$
\thanks{$^{1}$AUTOLAB at the University of California, Berkeley, USA.}%
\thanks{$^{2}$Google Research, USA.}%
\thanks{$^*$Work done while the author was an intern at Google.}%
\thanks{Correspondence to {\tt\small seita@berkeley.edu}}%
}
\begin{document}

\maketitle
\thispagestyle{empty}
\pagestyle{empty}

\begin{abstract}
Rearranging and manipulating deformable objects such as cables, fabrics, and bags is a long-standing challenge in robotic manipulation. The complex dynamics and high-dimensional configuration spaces of deformables, compared to rigid objects, make manipulation difficult not only for multi-step planning, but even for goal specification. Goals cannot be as easily specified as rigid object poses, and may involve complex relative spatial relations such as ``place the item inside the bag." In this work, we develop a suite of simulated benchmarks with 1D, 2D, and 3D deformable structures, including tasks that involve image-based goal-conditioning and multi-step deformable manipulation.  We propose embedding goal-conditioning into Transporter Networks, a recently proposed model architecture for learning robotic manipulation that rearranges deep features to infer displacements that can represent pick and place actions. \hl{In simulation and in physical experiments,} we demonstrate that goal-conditioned Transporter Networks enable agents to manipulate deformable structures into flexibly specified configurations without test-time visual anchors for target locations. We also significantly extend prior results using Transporter Networks for manipulating deformable objects by testing on tasks with 2D and 3D deformables. Supplementary material is available at \url{https://berkeleyautomation.github.io/bags/}.
\end{abstract}

\section{Introduction}\label{sec:intro}

Manipulating deformable objects is a long-standing challenge in robotics with a wide range of real-world applications. In contrast to rigid object manipulation, deformable object manipulation presents additional challenges due to more complex configuration spaces, dynamics, and sensing. 

While several prior works have made progress largely centered around tasks with 1D (\eg rope~\cite{nair_rope_2017}) or 2D (\eg fabric~\cite{seita_fabrics_2020}) deformable structures, little prior work has addressed generalizable vision-based methods for manipulating 3D deformable structures such as the task of: ``insert objects into a bag, and then carry the bag away''.  These types of tasks are especially challenging with diverse goal conditioning: the goal states of deformable objects are not easily specified, for example, by compact pose representations.

In this work, we propose a new suite of benchmark tasks, called \textit{DeformableRavens}, to test manipulation of cables, fabrics, and bags spanning 1D, 2D, and 3D deformables. For several tasks in the benchmark, we propose to tackle them using novel goal-conditioned variants of Transporter Network~\cite{zeng_transporters_2020} architectures.  
Our experiments also significantly extend results of using Transporter Networks for deformable manipulation tasks --- while~\cite{zeng_transporters_2020} demonstrated one 1D deformable task, we show results on 12 tasks, including those with fabrics and bags.


\begin{figure}[t]
\center
\includegraphics[width=0.49\textwidth]{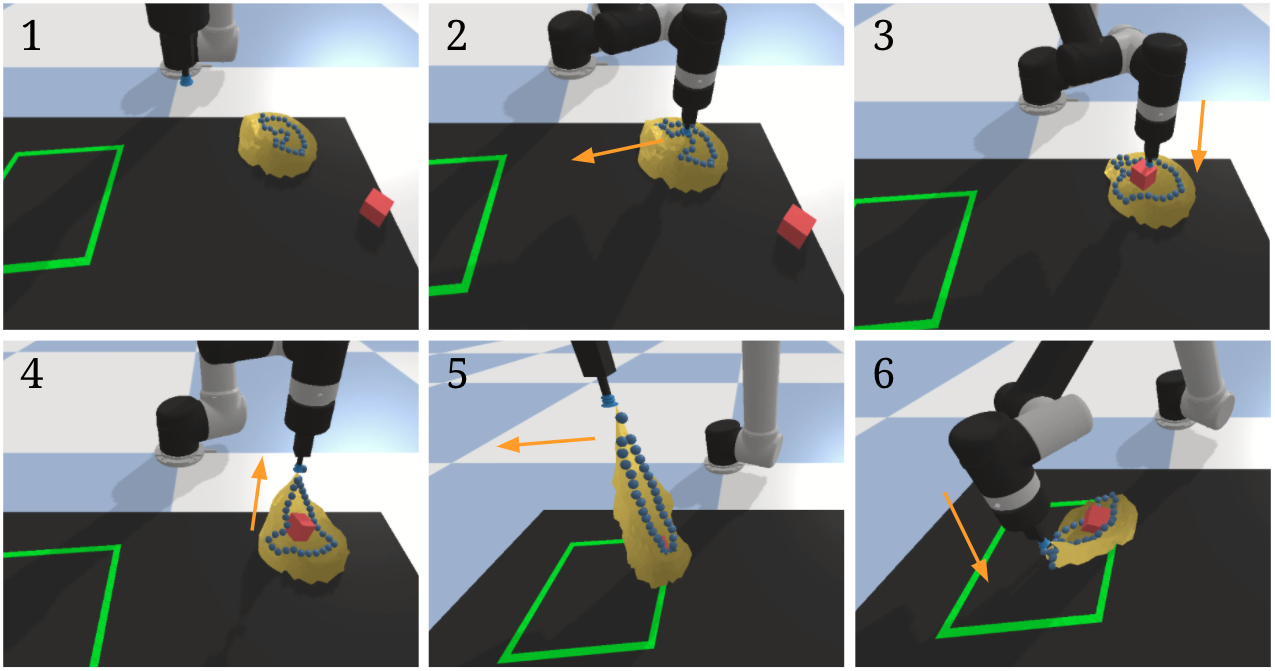}
\caption{
Example of a trained Transporter Network policy in action on the \emph{bag-items-1} task (see Table~\ref{tab:tasks}). The setup involves a simulated UR5 robot arm, a green target zone, a drawstring-style bag, and a red cube. The starting configuration is shown in the top left frame. The robot, with suction cup, must grasp and sufficiently open the bag to insert a cube, then (bottom row) picks and pulls upwards, enclosing the cube. The robot concludes by bringing the bag with the cube in the target zone. We overlay arrows to indicate the movement of the robot's arm just before a given frame.
}
\vspace*{-15pt}
\label{fig:teaser}
\end{figure}

The main contributions of this paper are: (i) an open-source simulated benchmark, \textit{DeformableRavens}, with 12 tasks manipulating 1D, 2D, and 3D deformable objects to help accelerate research progress in robotic manipulation of deformables, (ii) end-to-end goal-conditioned Transporter Network architectures that learn vision-based multi-step manipulation of deformable 1D, 2D, and 3D structures, and (iii) experiments demonstrating that the proposed vision-based architectures are competitive with or superior to baselines that consume ground truth simulated pose and vertex information. We also discuss the shortcomings of the system, which point to interesting areas for future research. The project website contains supplementary material, including the appendix, code, data, and videos.


\begin{figure*}[t]
\center
\includegraphics[width=1.00\textwidth]{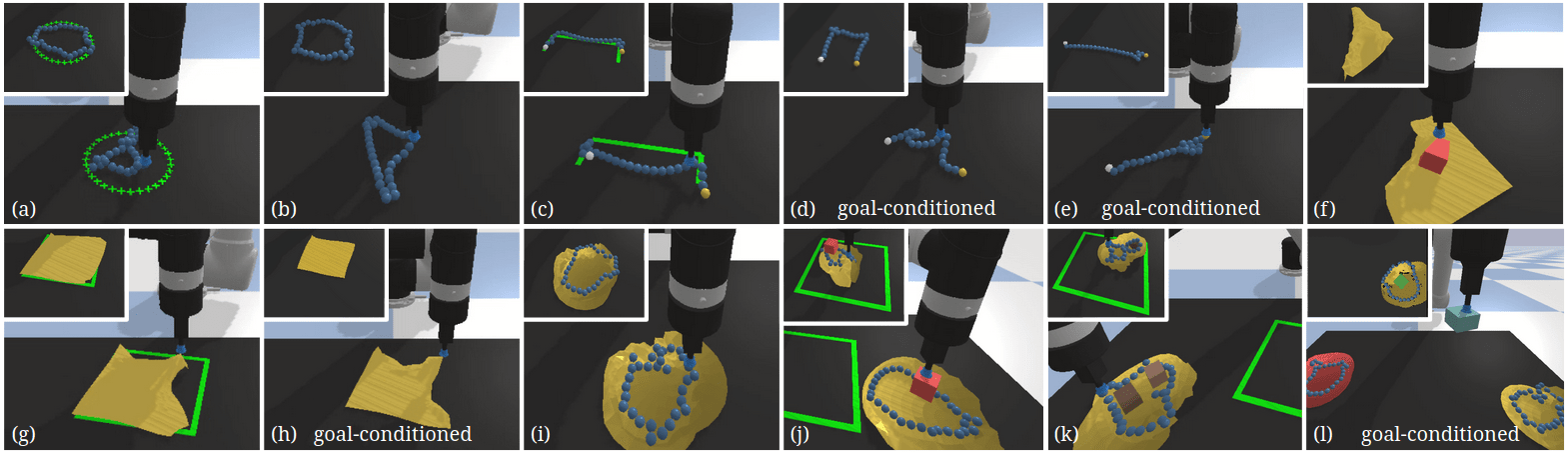} 
\caption{
The 12 tasks in the proposed \emph{DeformableRavens} benchmark (see Table~\ref{tab:tasks}) with suction cup gripper and deformable objects. \textbf{Top row}: (a) cable-ring, (b) cable-ring-notarget, (c) cable-shape, (d) cable-shape-notarget, (e) cable-line-notarget, (f) fabric-cover. \textbf{Bottom row}: (g) fabric-flat, (h) fabric-flat-notarget, (i) bag-alone-open, (j) bag-items-1, (k) bag-items-2, (l) bag-color-goal. Of the tasks shown, 4 are ``goal-conditioned,'' \ie cable-shape-notarget, cable-line-notarget, fabric-flat-notarget, bag-color-goal. These use a separate goal image $\bo_g$ for each episode to specify a success configuration. The other 8 tasks do not use goal images. Examples of successful configurations for each task are shown with overlaid (cropped) images to the top left.
}
\vspace*{-15pt}
\label{fig:tasks}
\end{figure*}

\section{Related Work}\label{sec:rw}


\subsection{Deformable Object Manipulation}

In this work, deformables~\cite{manip_deformable_survey_2018} refers to 1D structures such as ropes and cables (we use the terms interchangeably), 2D structures such as fabrics, and 3D structures such as bags.

Rope has been extensively explored in early robotics literature, such as for knot-tying~\cite{case_study_knots_1991,knot_planning_2003}. In recent years, learning-based approaches for manipulation have grown in popularity to improve the generalization of knot-tying and also to manipulate rope towards general target configurations~\cite{schulman_isrr_2013,nair_rope_2017,ZSVI_2018,priya_2020,state_estimation_LDO,wang_visual_planning_2019}. The approach we propose is focused primarily in manipulating cables to target locations, which may be specified by a target zone or a target image, and does not require modeling rope physics.

Much research on robotic fabric manipulation~\cite{grasp_centered_survey_2019,manip_deformable_survey_2018,corona_2018} uses bilateral robots and gravity to expose corners for ease of manipulation, which enables iteratively re-grasping the fabric's lowest hanging point~\cite{osawa_2007,kita_2009_iros,kita_2009_icra}. Subsequent work~\cite{maitin2010cloth,cusumano2011bringing,unfolding_rf_2014}, generalized to a wider variety of initial configurations of new fabrics. Other approaches focus on fabric smoothing.
For example, Sun~et~al.~\cite{heuristic_wrinkles_2014,cloth_icra_2015} attempt to detect and then pull perpendicular to the largest wrinkle. Another task of interest is wrapping rigid objects with fabric, which bridges the gap between manipulating 2D and 3D structures. Hayashi~et~al.~\cite{planning_wrapping_fabric_2017,fabric_wrapping_2014} present an approach for wrapping fabric around a cylinder using a bimanual robot.



Work on 3D deformable structures such as bags has been limited due to the complexities of manipulating bags. Early work focused on mechanical design of robots suitable for grasping~\cite{grasping_sacks_2005} or unloading~\cite{unloading_sacks_2008} sacks. Some work assumes that a sturdy, brown bag is already open for item insertion, as with a grocery checkout robot~\cite{checkout_robot_2011}, or uses reinforcement learning for zipping bags~\cite{contour_ziplock_2018} in constrained setups. 

These works generally focus on optimizing a specific problem in deformable object manipulation, often with task-specific algorithms or manipulators. In contrast, we propose a simulation setup with a single robot arm that applies to a wide range of tasks with 1D, 2D, and 3D deformables. In independent and concurrent work, Lin~et~al.~\cite{corl2020softgym}  propose \emph{SoftGym}, a set of environments with cables, fabrics, and liquids simulated using NVIDIA FleX, and use it to benchmark standard reinforcement learning algorithms.

\subsection{Data-Driven Robot Manipulation}

Robot manipulation using learned, data-driven techniques has become a highly effective paradigm for robot manipulation, as exemplified in the pioneering work of Levine~et~al.~\cite{levine_finn_2016}, Mahler~et~al.~\cite{mahler2019learning}, Florence~et~al.~\cite{dense_object_nets_2018}, Kalashnikov~et~al.~\cite{QT-Opt}, and others. Tools used often involve either Imitation Learning (IL)~\cite{argall2009survey} or Reinforcement Learning (RL)~\cite{Sutton_2018}, and in recent years, such techniques have been applied for manipulation of deformables.

For example, IL is used with human~\cite{seita-bedmaking} and scripted~\cite{seita_fabrics_2020} demonstrators for fabric smoothing, while RL is applied for smoothing and folding in~\cite{sim2real_deform_2018,rishabh_2020,lerrel_2020}. Using model-based RL,~\cite{visual_foresight_2018} were able to train a video prediction model to enable robots to fold pants and fabric, and a similar approach was investigated further in~\cite{fabric_vsf_2020} for model-based fabric manipulation. Some approaches combine IL and RL~\cite{balaguer2011combining}, and others use techniques such as latent space planning~\cite{latent_space_roadmap_2020,yan_fabrics_latent_2020} or using dense object descriptors~\cite{descriptors_fabrics_2020}. In this work, we use IL.

As described in Section~\ref{sec:PS}, the deep architecture we propose for image-based manipulation uses Fully Convolutional Neural Networks (FCNs)~\cite{fcn_2015} to produce per-pixel scores in an image, where each pixel corresponds to an action. This technique~\cite{thesis_andyzeng} has shown promising results for robotic warehouse order picking~\cite{zeng_icra_2018}, pushing and grasping~\cite{zeng_iros_2018}, tossing objects~\cite{zeng_tossing_2019}, kit assembly~\cite{form2fit_2020}, and mobile manipulation~\cite{spatial_action_maps_2020}. We show that similar approaches may be effective for picking and placing of deformables. In independent and concurrent work, Lee~et~al.~\cite{folding_fabric_fcn_2020} show how an approach can be used for fabric folding.

\section{Background}\label{sec:PS}

We first describe the problem formulation, followed by background on Transporter Networks~\cite{zeng_transporters_2020}. Section~\ref{sec:goal-conditioned} then describes a novel goal-conditioned framework.

\subsection{Problem Formulation}\label{ssec:problem-formulation}

We formulate the problem of rearranging deformable objects as learning a policy $\pi$ that sequences pick and place actions $\ba_t \in \mathcal{A}$ with a robot from visual observations $\bo_t \in \mathcal{O}$:
\begin{equation}\label{eq:action}
\pi(\bo_t)\rightarrow{\ba_t}=(\mathcal{T}_\textrm{pick},\mathcal{T}_\textrm{place})\in{\mathcal{A}}
\end{equation}
where $\mathcal{T}_\textrm{pick}$ is the pose of the end effector when grasping part of an object, and $\mathcal{T}_\textrm{place}$ is the pose of the end effector when releasing the grasp.
Since deformable objects may require more than a single pick and place action to reach a desired configuration, it is important that the policy $\pi$ learns from closed-loop visual feedback. Some rearrangement tasks~\cite{rearrangement_2020} may also be specified by a target goal observation $\bo_g$, in which case we refer to the policy as goal-conditioned: 
\begin{equation}\label{eq:action_goal}
\pi(\bo_t, \bo_g)\rightarrow{\ba_t}=(\mathcal{T}_\textrm{pick},\mathcal{T}_\textrm{place})\in{\mathcal{A}}.
\end{equation}
In this work, we consider tabletop manipulation tasks where both poses $\mathcal{T}_\textrm{pick}$ and $\mathcal{T}_\textrm{place}$ are defined in SE(2). Picking and placing positions are sampled from a fixed-height planar surface, while rotations are defined around the z-axis (aligned with gravity direction). The $\mathcal{T}_\textrm{pick}$ and $\mathcal{T}_\textrm{place}$ both parameterize a motion primitive \cite{frazzoli2005maneuver} that controls the end effector to approach $\mathcal{T}_\textrm{pick}$ until contact is detected,
activates the end effector to execute a grasp, moves upward to a fixed z-height, approaches $\mathcal{T}_\textrm{place}$, and lowers the end effector until contact is detected, then releases the grasp.
While this discrete-time planar action parameterization has its limitations, we find that it remains sufficiently expressive for a number of tabletop tasks involving manipulation of bags, in which gravity helps to induce a natural adhesive force that brings objects back towards the tabletop plane. 

To train the policy, we assume access to a small dataset of $N$ stochastic expert demonstrations $\mathcal{D} = \{\xi_i\}_{i=1}^N$, where each \emph{episode} $\xi_i$ of length $T_i$ (\ie the number of actions) consists of a sequence of observations and actions:
\begin{equation}\label{eq:tau}
\xi_i = (\bo_1, \ba_1, \bo_2, \ldots, 
\bo_{T_i}, \ba_{T_i}, \bo_{T_i + 1})
\end{equation}
used to supervise $\pi$, which may be goal-conditioned.

\begin{figure*}[t]
\center
\includegraphics[width=0.95\textwidth]{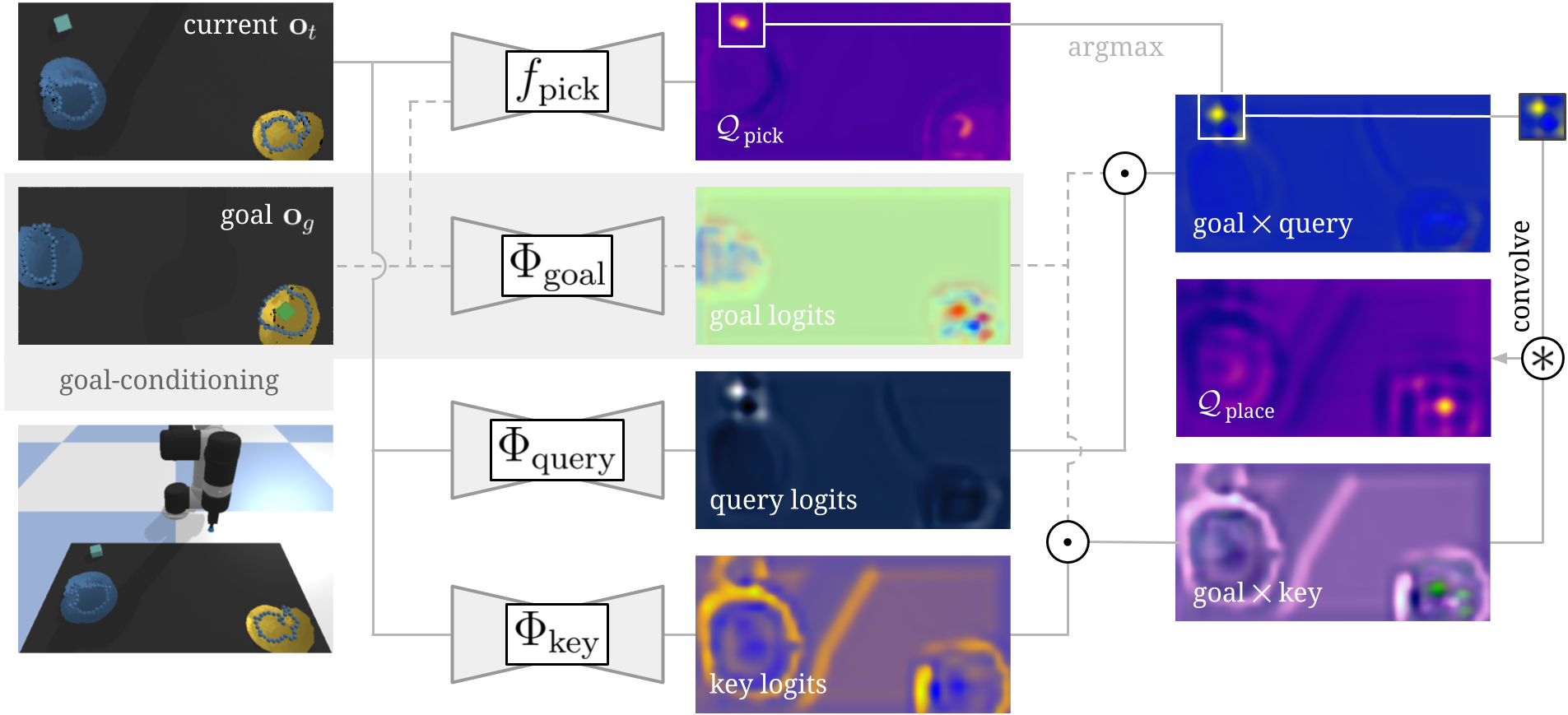}  
\caption{
The proposed \emph{Transporter-Goal-Split}, applied to an example on \emph{bag-color-goal}, where given the current image $\bo_t$ and goal $\bo_g$, the objective is to insert the block in the correct, open bag; the goal $\bo_g$ specifies that the yellow bag (with the block) is the target, not the blue bag. Transporter Networks use three Fully Convolutional Networks (FCNs), $f_\textrm{pick}$ for the attention module, and $\Phi_\textrm{query}$ and $\Phi_\textrm{key}$ for the transport module. \emph{Transporter-Goal-Split} considers a separate goal image $\bo_g$ and passes that through a new FCN, $\Phi_\textrm{goal}$. These are then combined with other deep features (labeled as ``logits'') through element-wise multiplication. Finally, the standard cross-convolution operator is applied. \emph{Transporter-Goal-Stack} dispenses with $\Phi_\textrm{goal}$ and instead stacks $\bo_t$ and $\bo_g$ as one image, and does not use $\Phi_\textrm{goal}$). All tested Transporter architectures produce $\mathcal{Q}_{\rm pick}$ and $\mathcal{Q}_{\rm place}$, which are each $320{\times}160$ dimensional heat maps colored so that darker pixels are low values and lighter pixels are high. The largest value in a heat map represents the \emph{pixel} to pick or place.
}
\vspace*{-10pt}
\label{fig:transporter-goal}
\end{figure*}

\subsection{Background: Transporter Networks}\label{ssec:vanilla-transporter}

\emph{Transporter Networks}~\cite{zeng_transporters_2020} is a model architecture for manipulation that learns to rearrange objects by (i) attending to a local region of interest, then (ii) predicting its target spatial displacement by cross-correlating its dense deep visual features over the scene. It can parameterize actions for pick and place, and has shown to work well for variety of tabletop manipulation tasks including stacking, sweeping, and kit assembly~\cite{form2fit_2020}. 

Transporter Networks consist of 3 Fully Convolutional Networks (FCNs). The first FCN $f_\textrm{pick}$ takes as input the visual observation $\bo_t$, and outputs a dense per-pixel prediction of action-values $\mathcal{Q}_\textrm{pick}$ that correlate with picking success: $\mathcal{T}_\textrm{pick} = \argmax_{(u, v)} \ \mathcal{Q}_\textrm{pick}((u,v)|\mathbf{o}_t)$ where each pixel $(u, v)$ corresponds to a picking action at that location via camera-to-robot calibration. The second FCN $\Phi_\textrm{key}$ also takes as input $\bo_t$, while the third FCN $\Phi_\textrm{query}$ takes as input a partial crop $\bo_t[\mathcal{T}_\textrm{pick}]$ from $\bo_t$ centered on $\mathcal{T}_\textrm{pick}$. Both the second and third FCNs output dense feature embeddings, which are then cross-correlated with each other to output a dense per-pixel prediction of action-values $\mathcal{Q}_\textrm{place}$ that correlate with placing success: $\mathcal{T}_\textrm{place} = \argmax_{\{\Delta\tau_i\}} \ \mathcal{Q}_\textrm{place}(\Delta\tau_i|\mathbf{o}_t,\mathcal{T}_\textrm{pick})$ where
\begin{equation}
\mathcal{Q}_\textrm{place}(\Delta\tau|\mathbf{o}_t,\mathcal{T}_\textrm{pick}) = \Phi_\textrm{query}(\mathbf{o}_t[\mathcal{T}_\textrm{pick}])*\Phi_\textrm{key}(\mathbf{o}_t)[\Delta\tau]
\end{equation}
and $\Delta\tau$ covers the space of all possible placing poses.

A key aspect of Transporter Networks is that the visual observations $\bo_t$ must be spatially consistent, so that the 3D structure of the data is preserved under different visuo-spatial transforms $\Delta\tau$. In practice, this property also improves training data augmentation (since rotations and translations of the orthographic image appear as different configurations of objects in the scene). To leverage this, observations $\bo_t$ are top-down orthographic images of the tabletop scene, where each pixel represents a vertical column of 3D space.

\section{Goal-Conditioned Transporter Networks}\label{sec:goal-conditioned}

In some tasks, it is more natural to specify a success criteria by passing in a goal observation $\bo_g$ showing objects in a desired configuration. In this case, we assume $\bo_g$ is fixed and available as an extra input with the current observation $\bo_t$, and that $\bo_t$ and $\bo_g$ have the same pixel resolution.

\subsection{Goal-Conditioned Transporter Networks}\label{ssec:goal-conditioned}

We propose two goal-conditioned architectures based on Transporter Networks. The first, \emph{Transporter-Goal-Stack}, stacks the current $\bo_t$ and goal $\bo_g$ images channel-wise, then passes it as input through a standard Transporter Network.
The second, \emph{Transporter-Goal-Split}, separates processing of the goal image $\bo_g$ through a fourth FCN $\Phi_\textrm{goal}$ to output dense features that are combined with features from the query $\Phi_\textrm{query}$ and key $\Phi_\textrm{key}$ networks using the Hadamard product:
\begin{equation}
\psi_\textrm{query}(\mathbf{o}_t) = \Phi_\textrm{query}(\mathbf{o}_t)\odot\Phi_\textrm{goal}(\mathbf{o}_t)
\end{equation}
\begin{equation}
\psi_\textrm{key}(\mathbf{o}_t) = \Phi_\textrm{key}(\mathbf{o}_t)\odot\Phi_\textrm{goal}(\mathbf{o}_t)
\end{equation}
\begin{equation}
\mathcal{Q}_\textrm{place}(\Delta\tau|\mathbf{o}_t,\mathcal{T}_\textrm{pick}) = \psi_\textrm{query}(\mathbf{o}_t)[\mathcal{T}_\textrm{pick}]*\psi_\textrm{key}(\mathbf{o}_t)[\Delta\tau]
\end{equation} We hypothesize that this separation is beneficial for learning, since convolutional kernels may otherwise struggle to disambiguate which channels correspond to $\bo_t$ or $\bo_g$.
Figure~\ref{fig:transporter-goal} shows the architecture of \emph{Transporter-Goal-Split}.
For consistency, all Transporter modules $\Phi_\textrm{query}$, $\Phi_\textrm{key}$, and $\Phi_\textrm{goal}$ employ a 43-layer, 9.9M parameter FCN~\cite{fcn_2015} with residual connections~\cite{resnets_2016}, given that similar architectures have been used for FCN-based picking and placing~\cite{zeng_icra_2018,zeng_transporters_2020}.

A key aspect of the goal-conditioned model is that the spatial structure of the goal images are preserved in the architecture. This prior encourages the deep networks to learn features that infer pick and place actions anchored on visual correspondences between objects in the current and goal images. This is useful for manipulating deformable objects, where changes between current and desired configurations are better captured with dense correspondences \cite{descriptors_fabrics_2020,dense_object_nets_2018}.

\subsection{Training Details for Goal Conditioned Models}

To train both goal-conditioned architectures, we use an approach based on Hindsight Experience Replay~\cite{HER_2017}. For each task (Table~\ref{tab:tasks}), we assume a dataset of $N$ demonstrations $\mathcal{D} = \{\xi_i\}_{i=1}^N$, where each episode $\xi_i$ (see Eq.~\ref{eq:tau}) is of length $T_i$ with final observation $\bo_g = \bo_{T_i + 1}$.
In standard Transporter Networks, a single sample $(\bo_k, \ba_k)$ containing the observation and the resulting action at time $k$, is drawn at random from $\mathcal{D}$. For the goal-conditioned Transporter Networks, we use the same procedure to get $(\bo_k, \ba_k)$, then additionally use the corresponding observation $\mathbf{o}_g$ after the last action from the demonstration episode containing $\bo_k$, to get training sample $(\bo_k, \ba_k, \bo_g)$. 

As described in Section~\ref{ssec:vanilla-transporter}, because observations $\bo_t$ are top-down orthographic images and are spatially consistent with rotations and translations, this enables data augmentation by randomizing a rotation and translation for each training image. To handle the goal-conditioned case, both $\bo_t$ and $\bo_g$ are augmented using the same random rotation and translation, to ensure consistency.

\section{Simulator and Tasks}\label{sec:simulator-tasks}


\begin{table}
  \setlength\tabcolsep{3.0pt}
  \centering
  \caption{
  \textbf{DeformableRavens}. Tasks involve rearranging deformable objects (\eg cables, fabrics, and bags). Each comes with a scripted expert demonstrator that succeeds with high probability, except for the four bag tasks which are challenging; for these, we filter to use only successful episodes in training. Some require \emph{precise placing} to trigger a success. Tasks with a \emph{visible zone} will have a green target zone on the workspace to indicate where items should be placed (e.g., a square target zone that a fabric must cover); other \emph{goal-conditioned} tasks use a separate goal image to specify the success criteria for object rearrangement. See Figure~\ref{fig:tasks} for visualizations.
  }
  \footnotesize
  \begin{tabular}{@{}lrccc}
  \toprule
  Task                  & demos$\;$  & precise & visible & goal  \\
  (Max. Episode Length) & stats(\%)  & placing & zone    & cond. \\
  \midrule
  (a) cable-ring$^\mathsection$ (20)          &  99.1 & \xmark & \cmark & \xmark \\
  (b) cable-ring-notarget$^\mathsection$ (20) &  99.3 & \xmark & \xmark & \xmark \\
  (c) cable-shape$^*$ (20)                    &  98.8 & \cmark & \cmark & \xmark \\
  (d) cable-shape-notarget$^*$ (20)           &  99.1 & \cmark & \xmark & \cmark \\
  (e) cable-line-notarget$^*$ (20)            & 100.0 & \cmark & \xmark & \cmark \\
  \midrule
  (f) fabric-cover (2)                        &  97.0 & \xmark & \xmark & \xmark \\
  (g) fabric-flat$^{\dagger}$ (10)            &  98.3 & \cmark & \cmark & \xmark \\
  (h) fabric-flat-notarget$^{\dagger}$ (10)   &  97.4 & \cmark & \xmark & \cmark \\
  \midrule
  (i) bag-alone-open$^{\mathsection}$ (8)     &  60.2 & \xmark & \xmark & \xmark \\
  (j) bag-items-1 (8)                         &  41.7 & \xmark & \cmark & \xmark \\
  (k) bag-items-2 (9)                         &  32.5 & \xmark & \cmark & \xmark \\
  (l) bag-color-goal (8)                      &  89.1 & \xmark & \xmark & \cmark \\
  \bottomrule
  \end{tabular}
  \\$^\mathsection$evaluated based on maximizing the convex hull area of a ring.
  \\$^*$evaluated by the percentage of a cable within a target zone.
  \\$^\dagger$evaluated using fabric coverage, as in Seita~et~al.~\cite{seita_fabrics_2020,seita-bedmaking}.
  \vspace{-2.0em}
  \label{tab:tasks}
\end{table}

We evaluate the system on \emph{DeformableRavens}, a novel suite of simulated manipulation tasks involving cables, fabrics, and bags, using PyBullet~\cite{coumans2019} with an OpenAI gym~\cite{gym} interface. See Table~\ref{tab:tasks} and Figure~\ref{fig:tasks} for overviews.

\subsection{Deformable Objects (Soft Bodies) in PyBullet}\label{ssec:simulator}

Data-driven methods in robot learning often require substantial amounts of data, which can be expensive to obtain in the real-world~\cite{QT-Opt}.
While simulators have been used in robotics to help alleviate this difficulty, such as for locomotion~\cite{sim-to-real-locomotion} and rigid object manipulation~\cite{openai-dactyl}, many simulated benchmarks for manipulation focus on rigid objects, partially because of difficulties in simulation of deformables~\cite{manipulation_cloth_liu_2016}.

Motivated by these challenges, we provide support for deformable objects (called ``soft bodies'') in PyBullet~\cite{coumans2019}, a widely-used publicly available simulator for robotics research. While prior work with soft bodies in PyBullet~\cite{deep_dressing_2018,assistive_gym_2020,sim2real_deform_2018} use position-based dynamics solvers, we use new soft body physics simulation based on the Finite Element Method~\cite{bathe2006finite} with mass-springs and self-collisions among vertices~\cite{cloth-cloth-collisions}. Contact and friction constraints between soft bodies and multi bodies are solved in a unified constraint solver at the velocity level. Implicit damping of the velocity uses a Krylov style method and soft body contact and friction is based on Conjugate Gradient for symmetric positive definite systems and Conjugate Residual for indefinite systems~\cite{Verschoor2019}.

\begin{figure}[t]
\center
\includegraphics[width=0.47\textwidth]{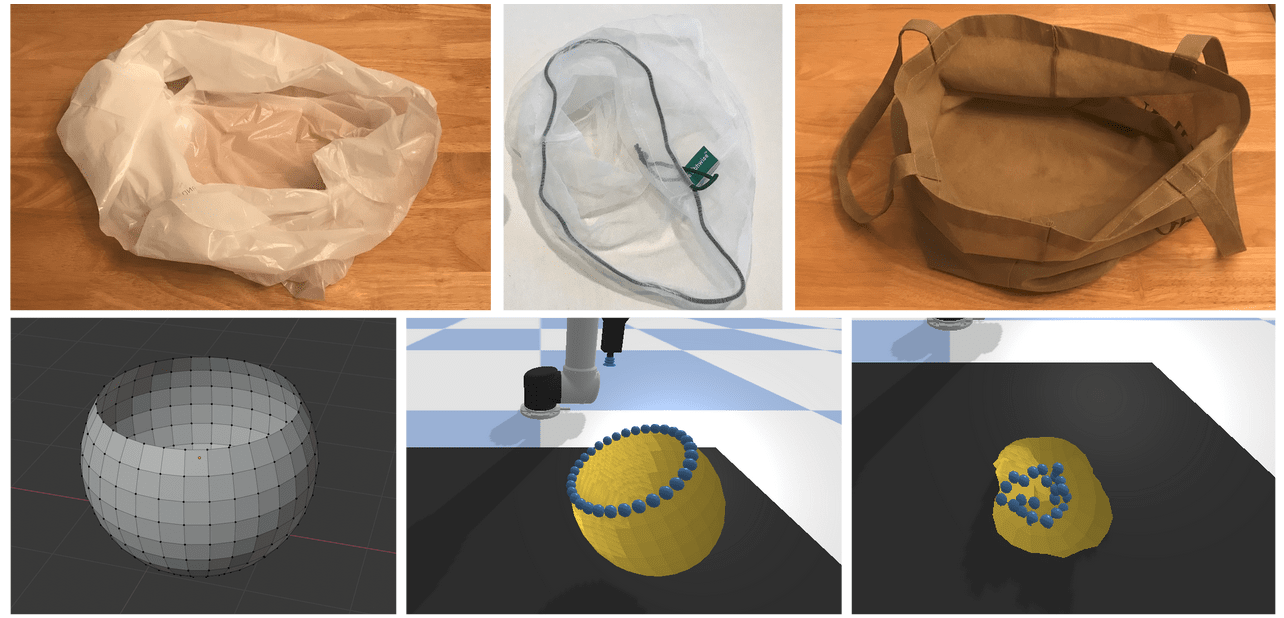} 
\caption{
\textbf{Top row}: examples of physical bags. The bags we use follow a design similar to the sack (top left) and drawstring (top middle). Bags with handles (\eg top right) or with more stiffness will be addressed in future work. \textbf{Bottom row}: to make bags, we create a sphere in Blender and remove vertices above a certain height (bottom left). We import the mesh into PyBullet and add a series of beads at the bag opening. For bag tasks, we initialize the bag by randomly sampling a starting pose (bottom middle), applying a small force, and allowing the bag to crumple (bottom right).
}
\vspace*{-10pt}
\label{fig:pybullet-sim}
\end{figure}

In simulation, cables consist of a sequence of rigid bodies (``beads''). Fabrics and bags are soft bodies, where each consists of a set of vertices. For fabrics and bags, we create a plane and a sphere object, respectively, using Blender~\cite{blender}. (For bags, we remove vertices to form an opening.) We then import the object mesh files into PyBullet. To improve physics stability of drawstring-style bagging behavior in PyBullet simulation, we additionally attach a ring of rigid beads at the bag's opening, using the same beads that form cables. See Figure~\ref{fig:pybullet-sim} for an overview. 

\subsection{Benchmark for Manipulating Deformable Objects}\label{ssec:benchmark}

We design 12 tasks with deformables, listed in Table~\ref{tab:tasks}. For consistency, each uses a standardized setup: a UR5 arm, a $0.5{\times}1$m tabletop workspace, and 3 calibrated RGB-D cameras diagonally overlooking the workspace, producing top-down observations $\bo_t \in \mathbb{R}^{320\times 160\times 6}$ of pixel resolution $320{\times}160$, where each pixel represents a $3.125{\times}3.125$mm vertical column of 3D space. Observations contain 3 RGB channels and 3 channel-wise depth values. The goal is to train a policy that executes a sequence of pick and place actions in the workspace to achieve an objective, learned from demonstrations with behavior cloning~\cite{Pomerleau_behavior_cloning}. Some tasks (\eg \emph{cable-shape-notarget} and \emph{fabric-flat-notarget}) are specified by goal images of a target scene configuration, while others are described solely by the distribution of demonstrations (\eg \emph{cable-ring-notarget} and \emph{bag-alone-open}).

The grasping motion primitive (introduced in Section~\ref{sec:PS}) is similar to the implementations in~\cite{fabric_vsf_2020,seita_fabrics_2020} and approximates a pinch-grasp on a deformable by indexing the nearest vertex that comes in contact with the end effector tip, and locking in its degrees of freedom to match that of the end effector using fixed constraints. Certain tasks are divided into different stages, such as \emph{bag-items-1} and \emph{bag-items-2} which involve opening a bag, inserting items inside, and lifting a bag. Depending on the task stage and the gripped item, actions lift the gripped object to a different hard-coded height. For example, pulling the bag upwards requires a higher vertical displacement than opening the bag.

\section{\hl{Simulation} Experiments}\label{sec:experiments}

We use scripted, stochastic demonstrator policies to get 1000 demonstrations (\ie episodes) per task, and train policies using 1, 10, 100, or all 1000 demonstrations. We provide a detailed overview of the data generation process in the Appendix. We test the following models:

\noindent\textbf{Transporter.} This model is directly from~\cite{zeng_transporters_2020}.

\noindent\textbf{Transporter-Goal-Split.} The Transport model that includes a separate goal module $\Phi_\textrm{goal}$ to process $\bo_g$.

\noindent\textbf{Transporter-Goal-Stack.} A model that stacks $\bo_t$ and $\bo_g$ channel-wise to form a 12-channel image, then passes it as input to standard Transporter Networks.

\noindent\textbf{GT-State MLP.} A model that uses ground truth pose information as observations $\bo_t$, and does not use images. It processes its input with a multi-layer perception (MLP) and directly regresses $\mathcal{T}_{\rm pick}$ and $\mathcal{T}_{\rm place}$.

\noindent\textbf{GT-State MLP 2-Step.} A baseline similar to the prior ground truth baseline, except it regresses in a two-step fashion, first regressing $\mathcal{T}_{\rm pick}$, and then concatenates this result again to the ground truth information to regress $\mathcal{T}_{\rm place}$.

We test \emph{Transporter} on non-goal conditioned tasks, whereas for those that use goals $\bo_g$, we test with \emph{Transporter-Goal-Split} and \emph{Transporter-Goal-Stack}. Ground truth baselines are tested in both settings, where in the goal-conditioned case, we concatenate the ground truth pose information in both the current and goal configurations to form a single input vector. Following~\cite{zeng_transporters_2020}, to handle multi-modality of picking and placing distributions, the models output a mixture density~\cite{mixture_density_networks} represented by a 26-D multivariate Gaussian. During training, all models use the same data augmentation procedures for a fair comparison.

\noindent\textbf{Evaluation metrics.}
Episodes for \emph{cable-shape}, \emph{cable-shape-notarget}, and \emph{cable-line-notarget}, are successful if all cable beads have reached designated target poses.
For \emph{cable-ring}, \emph{cable-ring-notarget}, and \emph{bag-alone-open}, success is triggered when the convex hull area of the ring of beads (forming the cable or the bag opening) exceeds a threshold.
Episodes for \emph{fabric-cover} are successful if the fabric covers the cube.
\emph{Fabric-flat} and \emph{fabric-flat-notarget} are evaluated based on coverage~\cite{seita_fabrics_2020}.
\emph{Bag-items-1} and \emph{bag-items-2} consider a success when all blocks have been transported to the target zone with the bag.
Finally, \emph{bag-color-goal} evaluates success if the item has been inserted into the correct colored bag.

\begin{table*}[t]
  \setlength\tabcolsep{5.0pt}
  \caption{
  \textbf{Results.} Task success rate (mean \% over 60 test-time episodes in simulation of the best saved snapshot) vs. \# of demonstration episodes (1, 10, 100, or 1000) used in training. For the first eight tasks listed, we benchmark with Transporter Networks~\cite{zeng_transporters_2020} (``Transporter'') and two baselines that use ground-truth pose information instead of images as input. The last row of four tasks tests goal-conditioned policies, and we test with proposed goal-conditioned Transporter Network architectures (see Section~\ref{ssec:goal-conditioned}), along with similar ground-truth baselines. We do not test the last four tasks with ``Transporter'' since the architecture does not support including an extra goal image as input. \hl{We bold the best-performing statistic for each experiment group.} Section~\ref{sec:experiments} and the Appendix contain more details. 
  }
  \centering
  \footnotesize
  \begin{tabular}{@{}lrrrrrrrrrrrrrrrr@{}} 
  \toprule
  & \multicolumn{4}{c}{cable-ring} & \multicolumn{4}{c}{cable-ring-notarget} & \multicolumn{4}{c}{cable-shape} & \multicolumn{4}{c}{fabric-cover}  \\
 \cmidrule(lr){2-5} \cmidrule(lr){6-9} \cmidrule(lr){10-13} \cmidrule(lr){14-17}
 Method & 1 & 10 & 100 & 1000 & 1 & 10 & 100 & 1000 & 1 & 10 & 100 & 1000 & 1 & 10 & 100 & 1000 \\
 \midrule
 GT-State MLP            & 0.0 & 0.0 & 0.0 & 0.0 & 0.0 & 1.7 & 3.3 & 5.0 & 0.4 & 0.8 & 1.0 & 0.5 & 3.3 & 25.0 & 18.3 & 21.7  \\
 GT-State MLP 2-Step     & 0.0 & 1.7 & 1.7 & 0.0 & 1.7 & 0.0 & 0.0 & 1.7 & 0.7 & 0.6 & 0.9 & 0.5 & 3.3 & 16.7 & 6.7 & 3.3  \\
 Transporter             & \textbf{16.7} & \textbf{50.0} & \textbf{55.0} & \textbf{68.3} & \textbf{15.0} & \textbf{68.3} & \textbf{73.3} & \textbf{70.0} & \textbf{75.6} & \textbf{80.6} & \textbf{90.1} & \textbf{86.5} & \textbf{85.0} & \textbf{100.0} & \textbf{100.0} & \textbf{100.0}  \\
 \midrule
 & \multicolumn{4}{c}{fabric-flat} & \multicolumn{4}{c}{bag-alone-open} & \multicolumn{4}{c}{bag-items-1} & \multicolumn{4}{c}{bag-items-2}  \\
 \cmidrule(lr){2-5} \cmidrule(lr){6-9} \cmidrule(lr){10-13} \cmidrule(lr){14-17}
 Method & 1 & 10 & 100 & 1000 & 1 & 10 & 100 & 1000 & 1 & 10 & 100 & 1000 & 1 & 10 & 100 & 1000 \\
 \midrule
 GT-State MLP            & 26.0 & 45.6 & 65.6 & 71.3 & 15.0 & 16.7 & 35.0 & 43.3 & 1.7 & 20.0 & 30.0 & 31.7 & 0.0 & 0.0 & 6.7 & 8.3  \\
 GT-State MLP 2-Step     & 21.8 & 30.9 & 45.5 & 41.7 & 11.7 & 15.0 & 18.3 & 26.7 & 0.0 & 8.3 & 28.3 & 31.7 & 0.0 & 1.7 & 6.7 & 11.7  \\
 Transporter             & \textbf{42.1} & \textbf{86.5} & \textbf{89.5} & \textbf{88.8} & \textbf{18.3} & \textbf{50.0} & \textbf{61.7} & \textbf{63.3} & \textbf{25.0} & \textbf{36.7} & \textbf{48.3} & \textbf{51.7} & \textbf{5.0}  & \textbf{30.0} & \textbf{41.7} & \textbf{46.7}  \\
 \midrule
 & \multicolumn{4}{c}{cable-line-notarget} & \multicolumn{4}{c}{cable-shape-notarget} & \multicolumn{4}{c}{fabric-flat-notarget} & \multicolumn{4}{c}{bag-color-goal}  \\
 \cmidrule(lr){2-5} \cmidrule(lr){6-9} \cmidrule(lr){10-13} \cmidrule(lr){14-17}
 Method & 1 & 10 & 100 & 1000 & 1 & 10 & 100 & 1000 & 1 & 10 & 100 & 1000 & 1 & 10 & 100 & 1000  \\
 \midrule
 GT-State MLP            & 9.4 & 45.3 & 73.8 & 75.7 & 9.9 & 44.9 & 62.7 & 65.9 & 18.2 & 50.2 & 63.7 & 64.1 & 0.0 & 0.5 & \textbf{10.0} & \textbf{19.8}  \\
 GT-State MLP 2-Step     & 9.5 & 32.0 & 56.9 & 71.5 & 9.6 & 44.0 & 54.3 & 52.7 & 16.0 & 43.0 & 50.5 & 62.9 & 0.0 & 1.4 & 5.5 & 0.0  \\
 Transporter-Goal-Split  & 76.2 & 94.7 & \textbf{100.0} & \textbf{100.0} & 45.5 & 75.0 & 87.8 & 93.7 & 38.2 & \textbf{76.8} & 86.8 & 88.1 & \textbf{12.2} & \textbf{29.8} & 0.0 & 10.0  \\
 Transporter-Goal-Stack  & \textbf{91.0} & \textbf{99.7} & \textbf{100.0} & \textbf{100.0} & \textbf{59.0} & \textbf{80.1} & \textbf{95.5} & \textbf{97.0} & \textbf{39.3} & 73.9 & \textbf{88.6} & \textbf{89.1} & 10.0 & 0.0 & 5.0 & 0.0  \\
  \bottomrule \vspace{-0.8em} \\
  \end{tabular}
  \label{tab:results}
  \vspace{-1.0em}
\end{table*}

\section{\hl{Simulation} Results}\label{sec:results}

We summarize results in Table~\ref{tab:results}. For each model type and demo count $N \in \{1, 10, 100, 1000\}$, we train 3 models randomly initialized with different TensorFlow~\cite{tensorflow2015-whitepaper} seeds, except for \emph{bag-color-goal}, where we report results from 1 model since the task is computationally expensive. Models are trained for 20K iterations, with a batch size of 1 for the three Transporter models and 128 for the two ground truth models (to strengthen the baseline). We save 10 snapshots throughout training at equally spaced intervals, and for each, we roll out 20 evaluation episodes. This gives 60 metrics (Section~\ref{sec:experiments}) for each snapshot, due to training 3 models. For each of the 10 snapshots and their associated metric, we average the 60 metrics and report the maximum over the 10 in Table~\ref{tab:results}. The Appendix has more extensive analysis.

\subsection{Non-Goal Conditioned Tasks}

In these tasks, \emph{Transporter} generally achieves orders of magnitude better sample efficiency than ground truth models. It performs reliably on \emph{fabric-cover} with 100\% success rates with just 10 demos (compared to 25\% or less for all ground truth models), and does well on \emph{cable-shape} ($\ge$~90.1\%) and \emph{fabric-flat} ($\ge$~86.5\%) given enough demos. On \emph{bag-alone-open}, \emph{Transporter} attains performance of 61.7\% and 63.3\% with 100 and 1000 demos, which is comparable to the scripted demonstrator performance of 60.2\% (1000 successes out of 1661) before data filtering. Similarly, on \emph{bag-items-1} and \emph{bag-items-2}, the best raw performance numbers (with 1000 demos) are 51.7\% and 46.7\%, which exceed demonstrator performance of 41.7\% and 32.5\% (see Table~\ref{tab:tasks}).

Qualitatively, the learned policies may generalize to new starting configurations from the same data distribution. Figure~\ref{fig:teaser} shows an example successful rollout on \emph{bag-items-1}, where the robot opens the bag, inserts the cube, picks and pulls the bag (which encloses the cube in it), then deposits the bag to the target zone. The project website contains videos.

\subsection{Goal Conditioned Tasks}

\hl{Across all 4 dataset sizes for \emph{cable-line-notarget}, \emph{cable-shape-notarget}, and \emph{fabric-flat-notarget}, both \emph{Transporter-Goal-Stack} and \emph{Transporter-Goal-Split} substantially outperform the two GT-State baselines. \emph{Transporter-Goal-Stack} achieves slightly higher performance among the cable-related tasks, though the gap narrows with more demonstrations in \emph{cable-line-notarget}, since both goal-conditioned Transporters each achieve 100\% success rates with 100 and 1000 demos. For \emph{fabric-flat-notarget}, the performance of both goal-conditioned Transporters is more evenly matched, while for \emph{bag-color-goal}, \emph{Transporter-Goal-Split} achieves higher success rates of 12.2\% and 29.8\% with 1 and 10 demos, respectively, though performance deteriorates with more demonstrations. Upon inspection of policy execution in \emph{bag-color-goal}, the goal-conditioned Transporters frequently perform ``destructive'' actions on the bag which significantly deform it, and makes it harder to recover an open bag later for item insertion. In future work, we will investigate methods for improving performance on \emph{bag-color-goal}.} 




\begin{figure}[t]
\center
\includegraphics[width=0.46\textwidth]{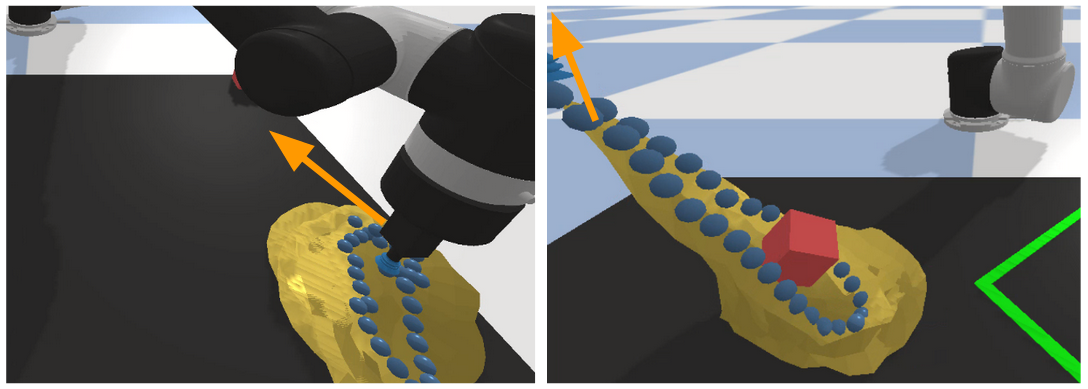}
\caption{
Failure cases we observe from trained Transporter policies on bag tasks. Left: in all bag tasks, a failure case may result from covering up the bag opening; these are hard for subsequent actions to recover from. Right: even if items are inserted into the bag, they may not be fully enclosed and can fall out when the bag is lifted. We overlay frames with orange arrows to indicate the direction of motion of the robot's end effector.
}
\vspace*{-10pt}
\label{fig:failure}
\end{figure}


\section{\hl{Physical Experiments and Results}}
\label{sec:physical-experiments-and-results}

\hl{We next validate experiments on physical hardware using a Franka Panda robot with a standard parallel-jaw gripper. The physical cable is a bead chain of length} \SI{45}{\centi\meter} \hl{when straightened. We use a flat foam rubber workspace with dimensions} \SI{60}{\centi\meter} by \SI{30}{\centi\meter}.

\subsection{Changes from Simulation}\label{ssec:heuristic}

\hl{Unlike in simulation, we cannot assume ``perfect'' grasping of deformable objects. As a heuristic, we take a local image crop centered at the picking point and compute the best fit tangent line of the cable. The robot opens its gripper finger tips in the direction perpendicular to that tangent line. Furthermore, to ease training, we use a binary segmentation mask as the input images, instead of color and depth.}

\subsection{Experiment Details and Protocol}

\begin{figure}[t]
\center
\includegraphics[width=0.44\textwidth]{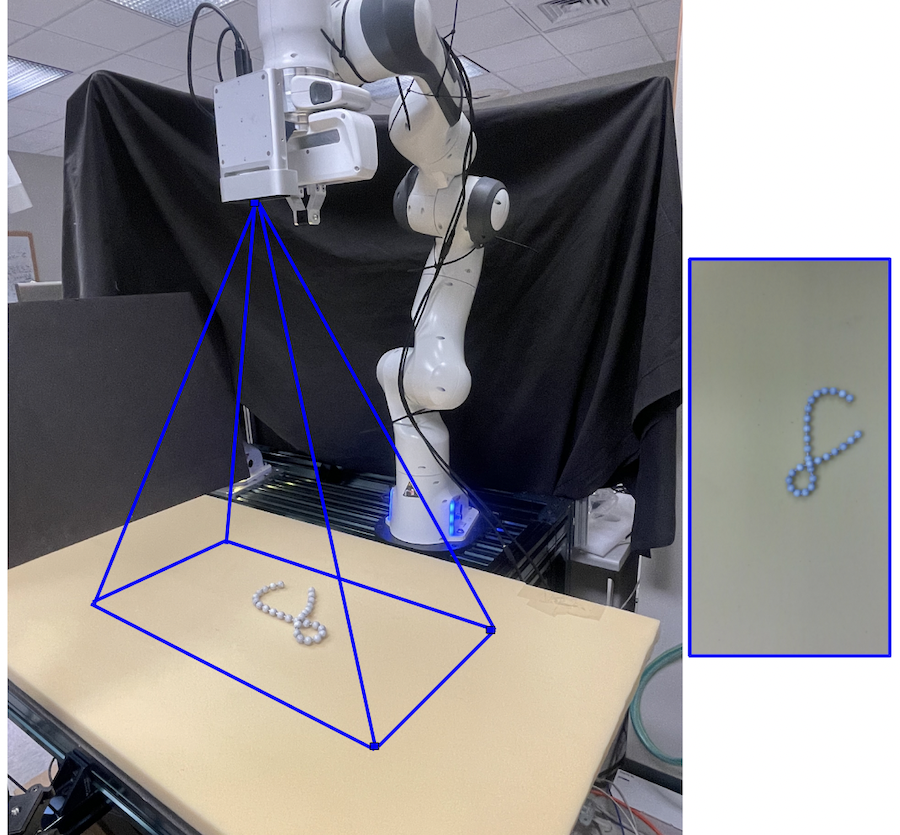}
\caption{
\hl{The physical setup from a third-person view, using a Franka robot with a hand-mounted camera. The blue contours show the field-of-view for the hand-mounted camera, and the inset image to the right shows the corresponding top-down RGB image.}
}
\vspace*{-5pt}
\label{fig:physical}
\end{figure}

\hl{We collect 30 human-provided demonstrations and use 24 for training, and monitor validation loss with the other 6. As with simulated experiments, we train \emph{Transporter-Goal-Split} on the training data for 20K iterations, where each iteration uses a batch size of 1 data example, rotated 24 times. We deploy the model snapshot with the lowest validation loss. We collect a separate set of 10 images of the cable on the workspace in various configurations to be used as goal images for testing.}

\hl{To interface with the robot, we use \texttt{frankapy}}~\cite{zhang2020modular}. \hl{The Franka uses a mounted Azure Kinect camera on its end-effector; it returns to a top-down home pose after each action to query updated images of the workspace. To initialize the setup, the human operator lightly tosses the cable on the workspace. We limit the number of robot actions to 10 per episode (i.e., rollout).}

\hl{As a quantitative evaluation metric, we report cable mask intersection over union (IoU). We compute this at each time from the current and goal cable mask images. We consider an episode as a success if the cable mask IoU reaches above a threshold within the 10-action limit. The threshold is 0.25, which corresponds to strong qualitative results but which is also forgiving of certain physical limitations. For example the robot will have inevitable imprecision when picking and placing due to slight errors in calibration.}

\subsection{Physical Results}


\hl{We execute 10 test-time episodes with the \emph{Transporter-Goal-Split} model, where each episode uses a different test-time goal image. The model succeeded in 7/10 of the episodes by exceeding the 0.25 cable mask IoU. It took an average of 7 pick-and-place actions per episode before achieving the success threshold.} 

\hl{Of the 3 failure cases, one occurred due to repeated grasping failures for a cable with significant self-overlap at the grasping point. This is partly due to the heuristic we use for rotating the gripper, described in Section}~\ref{ssec:heuristic}, \hl{which for cluttered cables, can create grasping angles that result in the robot ``pushing'' the cable towards the foam surface instead of grasping it. The other 2 failure episodes were due to inefficiencies in the policy, which performed repetitive back-and-forth picking and placing motions and did not reach the cable mask IoU threshold within the 10-action limit. See the project website for videos.}


\section{Limitations and Future Work}
\label{sec:limitations}

In this work, we present a new suite of tasks that evaluate end-to-end vision-based manipulation of deformable objects, spanning cables, fabrics, and bags. In addition, we propose goal-conditioned model architectures with Transporter Networks for rearranging objects to match goal configurations.

Manipulating deformable objects is challenging, and while the proposed method yields promising results, its limitations point to interesting directions for future work.
For example, Figure~\ref{fig:failure} shows several failure modes in the bag tasks.
The bag opening step is challenging because a single counterproductive action that visually occludes the bag opening makes it difficult to recover.
In addition, for \emph{bag-items-1} and \emph{bag-items-2}, items may fail to remain enclosed in the bag when lifted vertically.
\hl{This is in part due to the discrete-time planar parameterization of our pick and place action space.} 
Future work may investigate higher rates of control that learn recovery policies to react in real-time, or stateful systems that can track objects under visual occlusions, or bimanual manipulation policies, which may help prevent items from falling out of bags. 
\hl{Finally, while we report physical experiments using cables, in future work we will conduct physical experiments with fabrics and bags}.

\section*{Acknowledgments}

\small Daniel Seita is supported by the Graduate Fellowships for STEM Diversity. We thank Xuchen Han for assistance with deformables in PyBullet, and Julian Ibarz for helpful feedback on writing. \hl{We thank David Held for feedback on the physical experiments.}

\bibliographystyle{IEEEtranS}
\bibliography{example}

\normalsize
\cleardoublepage
\appendices

This appendix is structured as follows:

\begin{itemize}
    \item \hl{In Appendix}~\ref{app:changelog} \hl{we discuss the changes in various versions of the paper.}
    \item In Appendix~\ref{app:tasks} we describe the tasks in more detail.
    \item In Appendix~\ref{app:demonstrator-policy} we define the demonstrator policies.
    \item In Appendix~\ref{app:details-transporter} we provide more details of the various Transporter Network architectures.
    \item In Appendix~\ref{app:exp-details} we provide details of the experiment methodology.
    \item In Appendix~\ref{app:exp-results}, we present additional results.
\end{itemize}

More details, including videos of the learned policies and open-source code, are available on the project website.\footnote{\url{https://berkeleyautomation.github.io/bags/}}

\section{Paper Changelog}\label{app:changelog}

\hl{\textbf{Version 1} on arXiv was the initial publicly available edition of the paper. \textbf{Versions 2 and 3} made minor BibTeX and writing edits, and we ultimately submitted version 3 as the camera-ready for ICRA 2021. At that time,} the COVID-19 pandemic meant that it was infeasible for us to perform physical robotics experiments. \hl{Thus, for these paper versions, we ran all experiments using}  PyBullet simulation. 

\hl{In early 2023, we returned to this project to demonstrate how Goal-Conditioned Transporter Networks (GCTN) could be applied on physical hardware. In the newly-written Section}~\ref{sec:physical-experiments-and-results}, \hl{we report physical cable manipulation experiments based on the \emph{cable-line-notarget} task in simulation.}
\hl{In addition to the physical experiments, we fixed a bug in how we process the goal image data in both variants of Goal-Conditioned Transporter Networks (\emph{Transporter-Goal-Stack} and \emph{Transporter-Goal-Split}). This bug does not affect any of the other baselines, nor does it affect the vanilla Transporter networks. Fixing the bug results in improved performance for both GCTN versions across most tasks. For fairness, we also retrain the \emph{GT-State MLP} and \emph{GT-State MLP 2-Step}, which obtain similar results as earlier. Table}~\ref{tab:results} \hl{has the updated results.}
\hl{\textbf{Version 4} on arXiv thus has changes reflecting improved simulation performance of GCTNs and with new physical experiments.}



\section{Task Details}\label{app:tasks}

In \emph{DeformableRavens}, we propose 12 tasks, with details in Table~\ref{tab:tasks} and visualizations in Figure~\ref{fig:tasks}. Here, we elaborate upon the tasks and success criteria in more detail. In addition, we derive the size of the ground truth observation passed into the \emph{GT-State MLP} and \emph{GT-State MLP 2-Step} baseline models (see Section~\ref{sec:experiments}).

\subsection{Preliminaries}\label{app:preliminaries}

Each cable is formed from a set of either 24 or 32 rigid beads attached together. Each fabric consists of a grid of 100 vertices, and each bag consists of 353 vertices along with 32 rigid beads to form the bag opening (see Figure~\ref{fig:pybullet-sim}). The ground truth baselines account for every rigid object in a task by collecting the 2D coordinate position and a rotational angle about the $z$-axis, to get a single pose $(x,y,\theta)$. We account for fabrics by taking its 100 vertices and stacking all the 3D vertex positions together, to get a 300D vector. For bags, to reduce the ground truth state complexity and dimensionality, the baselines use ground truth pose information for the 32 rigid beads that form the bag opening, and do not use vertex information. The ground truth state also uses an additional 3D pose representing the displacement vector from the center of the bag's initial pose. Hence, a ground truth description of one bag is a vector of size $33 \times 3 = 99$. This reduced state representation to simplify the input to methods using ground truth information has been used in independent and similar work in benchmarking deformable object manipulation~\cite{corl2020softgym}.

The four bag tasks have different ``task stages,'' where the pull height and end-effector velocity vary based on the stage. All bag tasks have an initial stage where the robot opens up the bag if necessary. This stage has a low pull height and a slow gripper speed to increase physics stability. Then, \emph{bag-items-1}, \emph{bag-items-2}, and \emph{bag-color-goal} have stages where the robot must insert an item into the bag, and \emph{bag-items-1} and \emph{bag-items-2} additionally have a third stage after that where the robot must transport the bag to a target zone. In these later stages, the pull vector must be high enough so that either a rigid item can be successfully inserted into the bag opening, or that the bag can be lifted in midair. For test-time deployment, these task stages are determined based on a segmented image which determines the object corresponding to a pick point.

\subsection{DeformableRavens Task Details}

\textbf{(a) cable-ring} has a cable with 32 beads attached together in a ring. The robot must open the cable towards a target zone specified by a ring, which contains the same area as the cable with maximized convex hull area. The termination condition is based on whether the area of the convex hull of the cable is above a threshold. Size of ground truth state: with 32 beads and 32 targets (one per bead), each consisting of a 3D description, the state vector is of size $(32+32) \times 3 = 192$.

\textbf{(b) cable-ring-notarget} is the same as \emph{cable-ring}, except the target zone is not visible. Size of ground truth state: with 32 beads, the state vector is of size $32 \times 3 = 96$.

\textbf{(c) cable-shape} has a cable with 24 beads attached together, with both endpoints free (\ie it is not a ring-like cable). The robot must adjust the cable towards a green target shape anchored on the workspace, specified by a series of 2, 3, or 4 line segments attached to each other. Success is based on the fraction of beads that end up within a distance threshold of any point on the target zone. Size of ground truth state: with 24 beads, and 24 targets (one per bead), plus one more pose to represent the center of the target shape, all of which are represented with 3D poses, the state vector is of size $(24+24+1) \times 3 = 147$.

\textbf{(d) cable-shape-notarget} is the same as \emph{cable-shape}, except the target zone is not visible on the workspace, but specified by a separate goal image of the beads in a desired configuration. In this case, we save the poses of the 24 beads corresponding to the target image, and stack that with the ground-truth state vector. Size of ground truth: with 24 beads for the current cable and 24 for the cable in the target image, the state vector is of size $(24+24) \times 3 = 144$.

\textbf{(e) cable-line-notarget} is a simpler version of \emph{cable-shape-notarget}, where the target image is specified with beads that form roughly a straight line. Size of ground truth state: as with \emph{cable-shape-notarget}, the state vector is of size $(24+24) \times 3 = 144$.

\textbf{(f) fabric-cover} starts with a flat fabric and a cube. The robot must place the cube onto the fabric, and then fold the fabric so that it covers the cube. Success is based on if the cube is not visible via a top-down object mask segmentation. Size of ground truth state: we use a 300D representation of fabrics, along with 3D pose information from the cube, resulting in a state vector of size $300 + 3 = 303$.

\textbf{(g) fabric-flat} inspired by Seita~et~al.~\cite{seita_fabrics_2020,seita-bedmaking} and Wu~et~al.~\cite{lerrel_2020}, the robot must manipulate a square fabric to move it towards a visible target zone of the same size as a flat fabric. The fabric starts flat, but does not fully cover the target zone.\footnote{We tried perturbing the starting fabric configuration to induce folds, but found that it was difficult to tune self-collisions among fabric vertices.} Success is based on if the robot exceeds a coverage threshold of 85\%, computed via top-down image segmentation. Size of ground truth state: we use a 300D representation of fabrics, along with a 3D pose for the target zone, so the state vector is of size $300 + 3 = 303$.

\textbf{(h) fabric-flat-notarget} is the same as \emph{fabric-flat}, except the target is specified with a separate image showing the fabric in the desired configuration. For ground truth information, we store the 300D representation of the fabric in the target zone, and stack that as input with the current fabric. Size of ground truth state: with 300D representations for the current and target fabrics, the state vector is of size $300+300 = 600$.

\textbf{(i) bag-alone-open} has a single bag with random forces applied to perturb the starting configuration. The objective is to expand the bag opening, which is measured via the convex hull area of the ring of beads that form the bag opening. Success is based on whether the convex hull area exceeds a threshold. Size of ground truth state: as reviewed in Appendix~\ref{app:preliminaries}, each bag is a 99D vector, and that is the size of the state dimension as there are no other items.

\textbf{(j) bag-items-1} contains one bag, one cube, and one target zone, with forces applied to perturb the initial bag state. The goal is to first open the bag (if necessary), insert the cube into the bag, then lift and transport the bag to the target zone. The task is successful only if the cube is entirely contained in the target zone, and part of the bag opening is also in the target zone. Size of ground truth state: as with \emph{bag-alone-open}, the bag is 99D, and we additionally consider the 3D cube pose and the 3D target zone pose, so the state dimension is $99 + 3 + 3 = 105$. 

\textbf{(k) bag-items-2} is a harder variant of \emph{bag-items-1}, where there exists two items that must be transported to the target zone, and where the two items can be of various shapes and colors. The sizes of the items are constrained so that they can be feasibly inserted within the bag. Size of ground truth state: it contains similar information as \emph{bag-items-1}, but has one more rigid item, so the state dimension is $99 + 3 + 3 + 3 = 108$. 

\textbf{(l) bag-color-goal} is a goal conditioned bag task with two bags and one rigid item. The two bags have the same size but different colors. The target image specifies two bags with one item inside it, and the objective is for the robot to put the item in the correct colored bag. Critically, the current state will have a bag of the same color as the target bag, and in the same spatial location. Size of ground truth state: each bag is 99D, and there is one 3D rigid item pose. Finally, each bag is augmented with a 3D vector specifying the RGB color value. The dimension for the \emph{current} state is $(99 + 99 + 3 + 3 + 3)= 207$. This is repeated for the \emph{goal} state, hence the final state dimension is $207 \times 2 = 414$.

\subsection{Additional Task: Block-Notarget}\label{ssec:additional-task}

\begin{figure}[t]
\center
\includegraphics[width=0.48\textwidth]{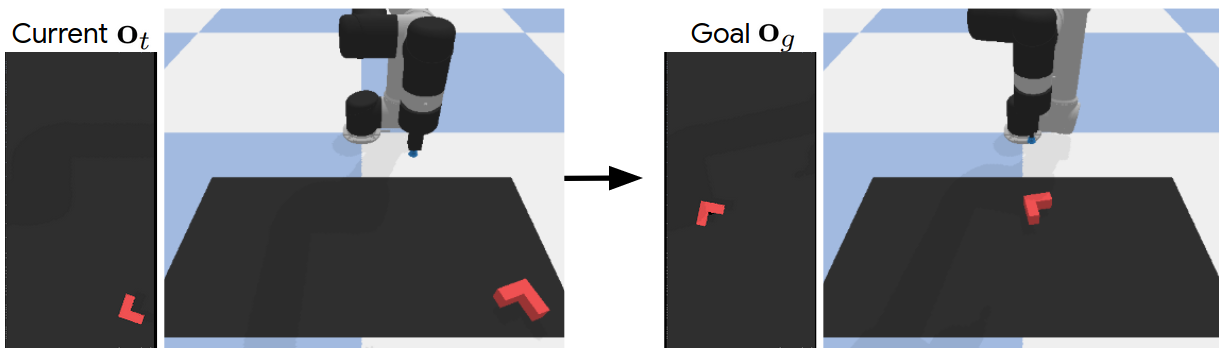}
\caption{
\textbf{block-notarget}. This task involves a red, rigid L-shaped block. At the current observation $\bo_t$, the block starts at some random location on the workspace, and the robot must use the same pick and place action formulation to get the block to a target pose specified in the goal image $\bo_g$.
}
\label{fig:block-notarget}
\end{figure}

In addition to tasks in \emph{DeformableRavens}, we create \textbf{block-notarget}, for fast prototyping of goal-conditioned Transporter Network architectures. This task is based on the insertion task from Zeng~et~al.~\cite{zeng_transporters_2020}, where the robot must pick, rotate, and precisely place an L-shaped block into a target receptacle. To adapt it to the goal-conditioned setting, we remove the the visible receptacle, and instead pass a separate goal image $\bo_g$ which contains the receptacle in the desired location. See Figure~\ref{fig:block-notarget} for a visual. The ground truth state vector is of size 6, due to having 3D pose information for the current block and target block.

\section{Demonstrator Data Policy}\label{app:demonstrator-policy}

\begin{table}[t]
  \setlength\tabcolsep{3.0pt}
  \centering
  \caption{
  Further details on the scripted demonstrator data performance on tasks in \emph{DeformableRavens}, along with the extra \emph{block-notarget} task. See Table~\ref{tab:tasks} in the main text. For each task and its 1000 demonstrator data episodes (which are filtered to only contain successes in the four bag tasks), we report the mean and median length. See Appendix~\ref{app:demonstrator-policy} for details. 
  }
  \footnotesize
  \begin{tabular}{@{}lrr}
  \toprule
  Task (Max Ep. Length) & Mean & Median \\
  \midrule
  cable-ring (20)           & $6.56 \pm 3.8$ & 5.0 \\
  cable-ring-notarget (20)  & $6.49 \pm 3.7$ & 5.0 \\
  cable-shape (20)          & $6.86 \pm 3.3$ & 6.0 \\
  cable-shape-notarget (20) & $6.94 \pm 3.4$ & 6.0 \\
  cable-line-notarget (20)  & $5.55 \pm 2.2$ & 6.0 \\
  \midrule                      
  fabric-cover (2)          & $2.00 \pm 0.0$ & 2.0 \\
  fabric-flat (10)          & $3.10 \pm 1.5$ & 3.0 \\
  fabric-flat-notarget (10) & $3.12 \pm 1.6$ & 3.0 \\
  \midrule                     
  bag-alone-open (10)       & $5.63 \pm 3.4$ & 5.0 \\
  bag-items-1 (8)           & $2.84 \pm 0.8$ & 3.0 \\
  bag-items-2 (9)           & $4.07 \pm 0.9$ & 4.0 \\
  bag-color-goal (8)        & $2.60 \pm 1.3$ & 2.0 \\
  \midrule
  block-notarget (2)        & $1.00 \pm 0.0$ & 1.0 \\
  \bottomrule
  \end{tabular}
  \vspace{-1.0em}
  \label{tab:demo_data_stats}
\end{table}

All tasks come with a scripted demonstrator. We briefly highlight how the demonstrator is scripted, and refer the reader to the open-source code for details.

For \emph{cable-shape}, \emph{cable-shape-notarget}, and \emph{cable-line-notarget}, the demonstrator computes, for each cable bead, its distance to the nearest point on the target zone. It performs a pick and place to correct the largest discrepancy. For \emph{fabric-cover}, the demonstrator places the cube onto the center of the flat fabric, then picks a corner at random and pulls it to the opposite corner of the fabric to cover the item. For \emph{fabric-flat} and \emph{fabric-flat-notarget}, we use a corner-pulling demonstrator as in~\cite{seita_fabrics_2020} by sequentially correcting for the fabric corner furthest from the target zone corner.

In \emph{cable-ring}, \emph{cable-ring-notarget}, and all four bag tasks, the demonstrator begins with a common sub-procedure. These tasks contain a set of attached beads that either form a ring or the bag's opening (for bag tasks). For each action in this stage, the demonstrator first computes a set of target positions on the workspace (forming a circle) corresponding to what a maximally spread out cable ring would look like. It then enumerates all valid bipartite graphs between the set of cable beads and the set of targets, and picks the one with minimum distance, and concludes with a pick and place to correct the largest bead-to-target discrepancy.

Once the area of the convex hull contained by the beads exceeds a tuned threshold, the task either ends (if \emph{cable-ring}, \emph{cable-ring-notarget}, or \emph{bag-alone-open}) or the demonstrator proceeds to inserting items (if \emph{bag-items-1}, \emph{bag-items-2}, or \emph{bag-color-goal}). For inserting items, the demonstrator picks at a valid item and then places it at a random spot within the bag opening, computed via OpenCV contour detection. Finally, for the bag moving stage in \emph{bag-items-1} and \emph{bag-items-2}, the demonstrator does a pick on a random visible bag bead, pulls with a hard-coded higher height delta, and ultimately places the bag at the center of the target zone.

We briefly highlight two aspects of the demonstrator:

\begin{enumerate}
\item Since the demonstrator has access to ground truth state information (\ie rigid object poses and deformable vertices), it performs the same on tasks that only differ based on the existence of a visible target zone on the workspace (\eg \emph{fabric-flat} versus \emph{fabric-flat-notarget}).

\item The demonstrator is stochastic (\ie behaviors are different between demonstrations). PyBullet provides built-in image segmentation code to find all pixels corresponding to an object. When the demonstrator picks an object, it samples uniformly at random among all pixels corresponding to the target object in the segmentation mask. The stochastic nature, along with randomly sampled object poses at the start of each new episode, ensures some level of dataset diversity.
\end{enumerate}

Since the models we test use behavior cloning, the performance of the learned agents is bounded by the data-generating policy. For the bag tasks, it is difficult to script a high-performing demonstrator given the complexity of bag simulation and manipulation. Therefore, we ignore unsuccessful episodes and run the demonstrator to generate as much data as needed until it gets 1000 successful demos. Due to stochasticity, we hypothesize that seeing only the successful demonstrations may be sufficient for policies to learn reasonable pick and place patterns from data. For \emph{bag-alone-open}, \emph{bag-items-1}, \emph{bag-items-2}, and \emph{bag-color-goal}, getting 1000 successful demonstrator episodes required running 1661, 2396, 3276, and 1110 times, respectively.

\section{Additional Transporter Network Details}\label{app:details-transporter}

For standard Transporter Networks~\cite{zeng_transporters_2020} (reviewed in Section~\ref{ssec:vanilla-transporter}), we swap the cropping order within the query network $\Phi_{\rm query}$, so that cropping happens after the observation image $\bo_t$ passes through the network. This enables subsequent features to use a larger receptive field.

For \emph{Transport-Goal-Split}, we demonstrate the forward pass for the transport module using Tensorflow-style~\cite{tensorflow2015-whitepaper} pseudo-code. This module includes $\Phi_{\rm key}$, $\Phi_{\rm query}$, and $\Phi_{\rm goal}$, but does not include the attention module $f_{\rm pick}$, which is used prior to the transport module forward pass.

\begin{python}[t]
import numpy as np
import tensorflow as tf
import tensorflow_addons as tfa

# Args:
#   p (pick point tuple, from `f_pick` net)
#   crop_size (64 in our experiments)
#   n_rots (usually 1 in our experiments)
#   input_data and goal_data

# Define FCNs for Transport module.
Phi_key   = FCN()
Phi_query = FCN()
Phi_goal  = FCN()

# input and goal images --> TF tensor.
in_tensor = tf.convert_to_tensor(input_data)
g_tensor  = tf.convert_to_tensor(goal_data)

# Get SE2 rotation vectors for cropping, centered
# at pick point. Assumes it's defined somewhere.
rvecs = get_se2(n_rots, p)

# Forward pass through three separate FCNs.
k_logits = Phi_key(in_tensor)
q_logits = Phi_query(in_tensor)
g_logits = Phi_goal(g_tensor)

# Combine.
g_x_k_logits = tf.multiply(g_logits, k_logits)
g_x_q_logits = tf.multiply(g_logits, q_logits)

# Crop the kernel_logits about the picking point,
# with the help of TensorFlow's image transforms.
crop = tf.identity(g_x_q_logits)
crop = tf.repeat(crop, repeats=n_rots, axis=0)
crop = tfa.image.transform(crop, rvecs)
kernel = crop[:,
              p[0]:(p[0] + crop_size),
              p[1]:(p[1] + crop_size),
              :]

# Cross-convolve!
output = tf.nn.convolution(g_x_k_logits, kernel)
return (1 / (crop_size**2)) * output
\end{python}

%
    
\section{Additional Experiment Details}\label{app:exp-details}

\begin{table}[t]
  \setlength\tabcolsep{3.0pt}
  \centering
  \caption{
  Some relevant hyperparameters of models tested and reported in Table~\ref{tab:results}. For brevity, we use ``Transporters'' here to collectively refer to Transporter, Transporter-Goal-Split, and Transporter-Goal-Stack. See Sections~\ref{sec:PS} and~\ref{sec:goal-conditioned} for background on notation for Transporter-based models. We also report values for the ground truth (GT) baseline models.
  }
  \footnotesize
  \begin{tabular}{@{}lc}
  \toprule
  Hyperparameter & Value  \\ 
  \midrule
  Adam Learning Rate ($f_{\rm pick}, \Phi_{\rm key}, \Phi_{\rm query}, \Phi_{\rm goal}$)     & 1e-4    \\
  Adam Learning Rate (GT)   & 2e-4   \\
  Batch Size (Transporters) & 1      \\
  Batch Size (GT)           & 128    \\
  Training Iterations (Transporters and GT) & 20,000 \\
  \bottomrule
  \end{tabular}
  \label{tab:hparams}
\end{table}

Zeng~et~al.~\cite{zeng_transporters_2020} test with several baseline models in addition to Transporters: Form2Fit~\cite{form2fit_2020}, a CNN followed by fully connected layers for actions~\cite{levine_finn_2016}, two ground-truth state baselines, and ablations to Transporter Networks. In an effort to alleviate compute, we only use the best versions of the Transporter Networks and Ground-Truth methods in~\cite{zeng_transporters_2020}, and we do not use the Form2Fit or CNN baselines, which exhibited significantly worse performance than Transporters. We show relevant hyperparameters, such as the Adam~\cite{adam2015} learning rate, in Table~\ref{tab:hparams}.

As described in Section~\ref{sec:results}, for each model type and demo count $N \in \{1, 10, 100, 1000\}$, we train for 20K iterations, where each iteration consists of one gradient update from one minibatch of data. We save snapshots every 2K iterations, resulting in 10 snapshots. Each snapshot is then loaded to run 20 episodes using held-out ``test set'' random seeds. We repeat this process by considering 3 separate training runs (per model type and demo count) to get more evaluation data points. The exception is with \emph{bag-color-goal}, where we train and load (\ie test) just 1 time (instead of 3) for each demo count, due to the computationally intensive nature of testing, which requires simulating environment steps with two bags. For the 100 and 1000 demonstration \emph{bag-color-goal} datasets, we additionally train \emph{Transporter-Goal-Split} and \emph{Transporter-Goal-Stack} for 40K iterations instead of 20K (while keeping the batch size at 1), since preliminary results showed that performance of the models notably improved at around 20K iterations.

\section{Additional Experiment Results}\label{app:exp-results}

\begin{figure*}[t]
\center
\includegraphics[width=1.00\textwidth]{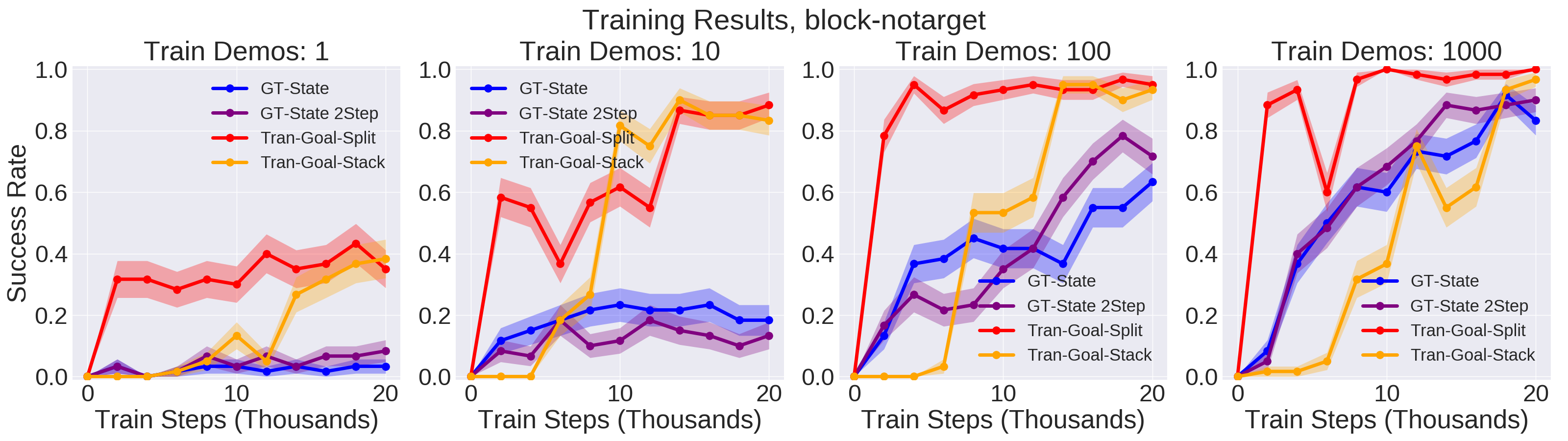}
\caption{
Results for models on the \emph{block-notarget} task. From left to right, we report models trained with 1, 10, 100, and 1000 demonstrations. All models are trained for 20K iterations, and perform test-time evaluation on iterations: 2K, 4K, 6K, 8K, 10K, 12K, 14K, 16K, 18K, 20K. Each data point represents the average of 60 test-time episode evaluations; we train 3 separate training runs, and run each of those snapshots for 20 test-time episodes at each iteration checkpoint. We also report performance at ``iteration 0'' with a randomly-initialized policy. Shaded regions indicate one standard error of the mean.
}
\label{fig:block-notarget-curves}
\end{figure*}

\begin{table}[t]
  \setlength\tabcolsep{3.0pt}
  \centering
  \caption{
  \textbf{Baseline Comparisons on Block Task}. Task success rate specifically for \emph{block-notarget}, described in Appendix~\ref{ssec:additional-task} and Figure~\ref{fig:block-notarget}. The table is formatted in a similar way as in Table~\ref{tab:results}.
  }
  \footnotesize
  \begin{tabular}{@{}lrrrr}
  \toprule
  & \multicolumn{4}{c}{block-notarget} \\
  \cmidrule(lr){2-5}
  Method & 1 & 10 & 100 & 1000 \\
  \midrule
  GT-State MLP           & 3.3 & 23.3 & 63.3 & 91.7 \\
  GT-State MLP 2-Step    & 8.3 & 18.3 & 78.3 & 90.0 \\
  Transporter-Goal-Stack & 38.3 & 90.0 & 95.0 & 96.7 \\
  Transporter-Goal-Split & 43.3 & 88.3 & 96.7 & 100.0 \\
  \toprule
  \end{tabular}
  \vspace{-0.5em}
  \label{tab:block-notarget}
\end{table}

We find in Table~\ref{tab:block-notarget} that \emph{Transporter-Goal-Stack} and \emph{Transporter-Goal-Split} are roughly two orders of magnitude more sample efficient on \emph{block-notarget}, with success rates from 10 demos (88.3\% and 90.0\% for the two variants) on par with success rates from 1000 demos for the ground truth models (90.0\% and 91.7\%). Figure~\ref{fig:block-notarget-curves} contain learning curves showing the performance of the models as a function of training iterations.

\begin{figure*}[t]
\center
\includegraphics[width=1.00\textwidth]{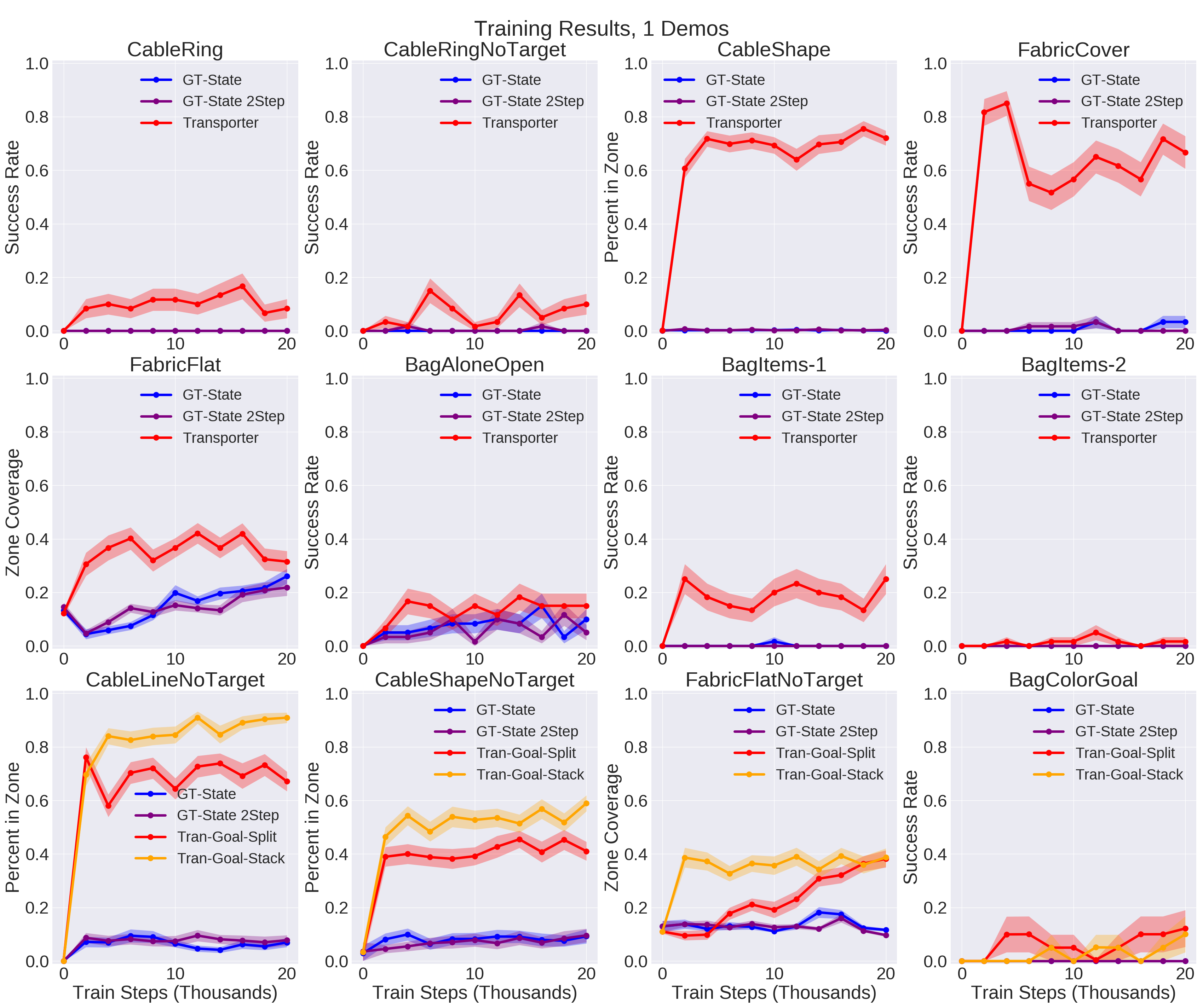}
\caption{
Results for various models, each trained on \textbf{1 demonstration}, with shaded regions indicating one standard error of the mean. We train for 20K iterations, and perform test-time evaluation on iterations: 2K, 4K, 6K, 8K, 10K, 12K, 14K, 16K, 18K, 20K. Each data point represents the average of 60 test-time episode evaluations; we train three separate training runs, and run each snapshot within those runs for 20 test-time episodes. Random seeds for test-time episodes are kept fixed among all methods so that evaluation is done on consistent starting states. We also report performance of policies at ``iteration 0'' with randomly-initialized parameters.
}
\label{fig:results-0001-demos}
\end{figure*}

\begin{figure*}[t]
\center
\includegraphics[width=1.00\textwidth]{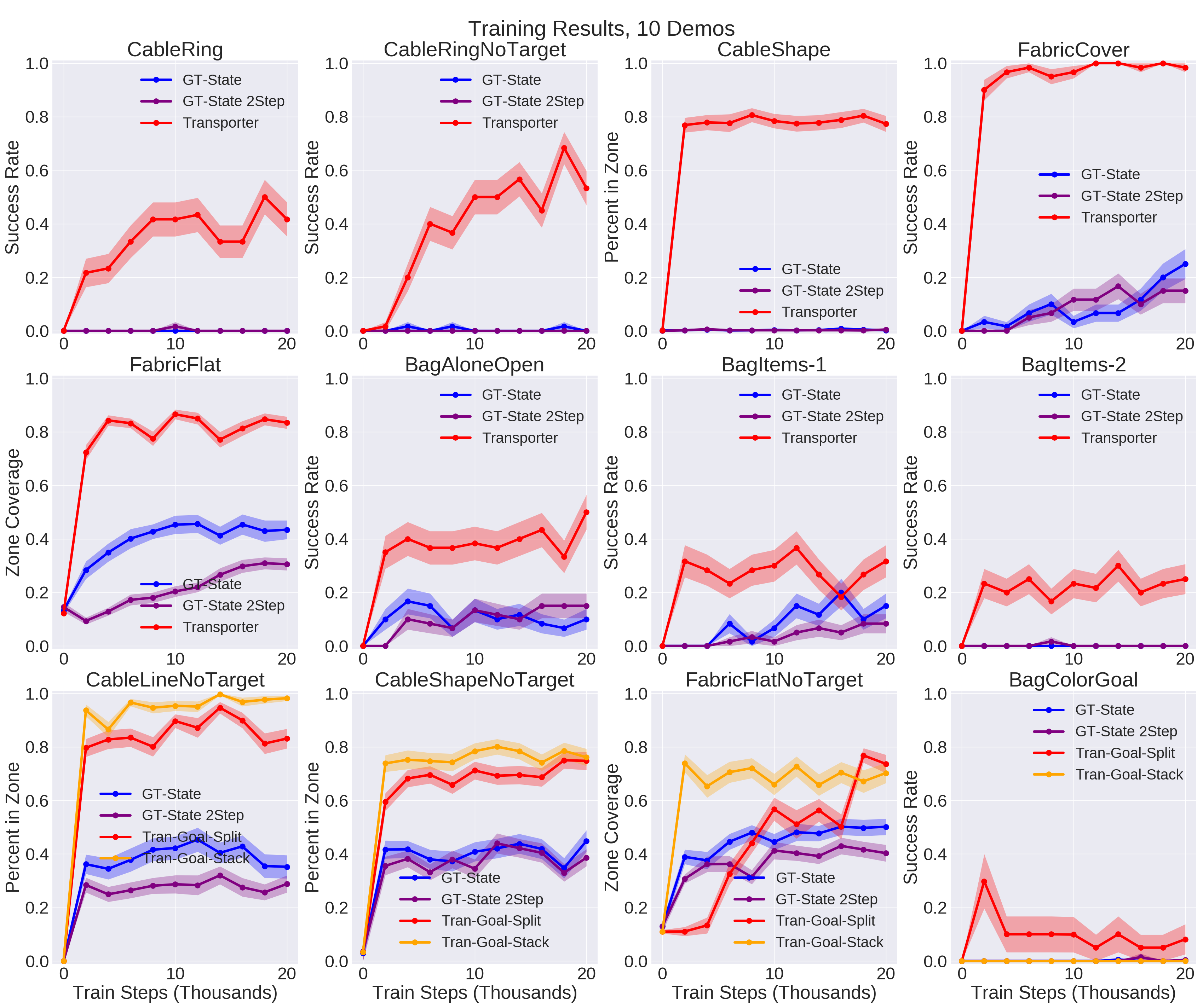}
\caption{
Results for various models, each trained on \textbf{10 demonstrations}. The results and plots are formatted following Figure~\ref{fig:results-0001-demos}.
}
\label{fig:results-0010-demos}
\end{figure*}

\begin{figure*}[t]
\center
\includegraphics[width=1.00\textwidth]{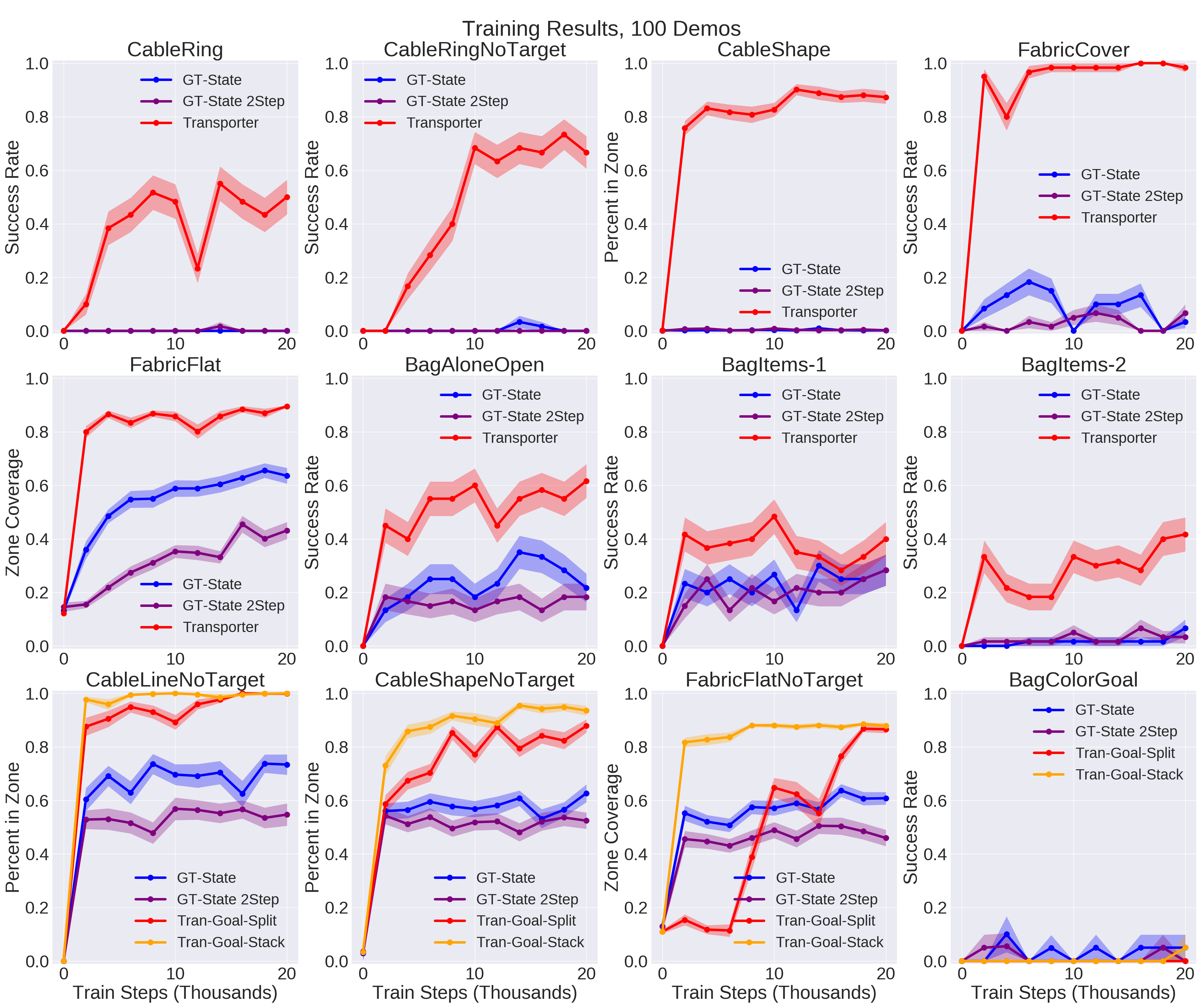}
\caption{
Results for various models, each trained on \textbf{100 demonstrations}. The results and plots are formatted following Figure~\ref{fig:results-0001-demos}. 
}
\label{fig:results-0100-demos}
\end{figure*}

\begin{figure*}[t]
\center
\includegraphics[width=1.00\textwidth]{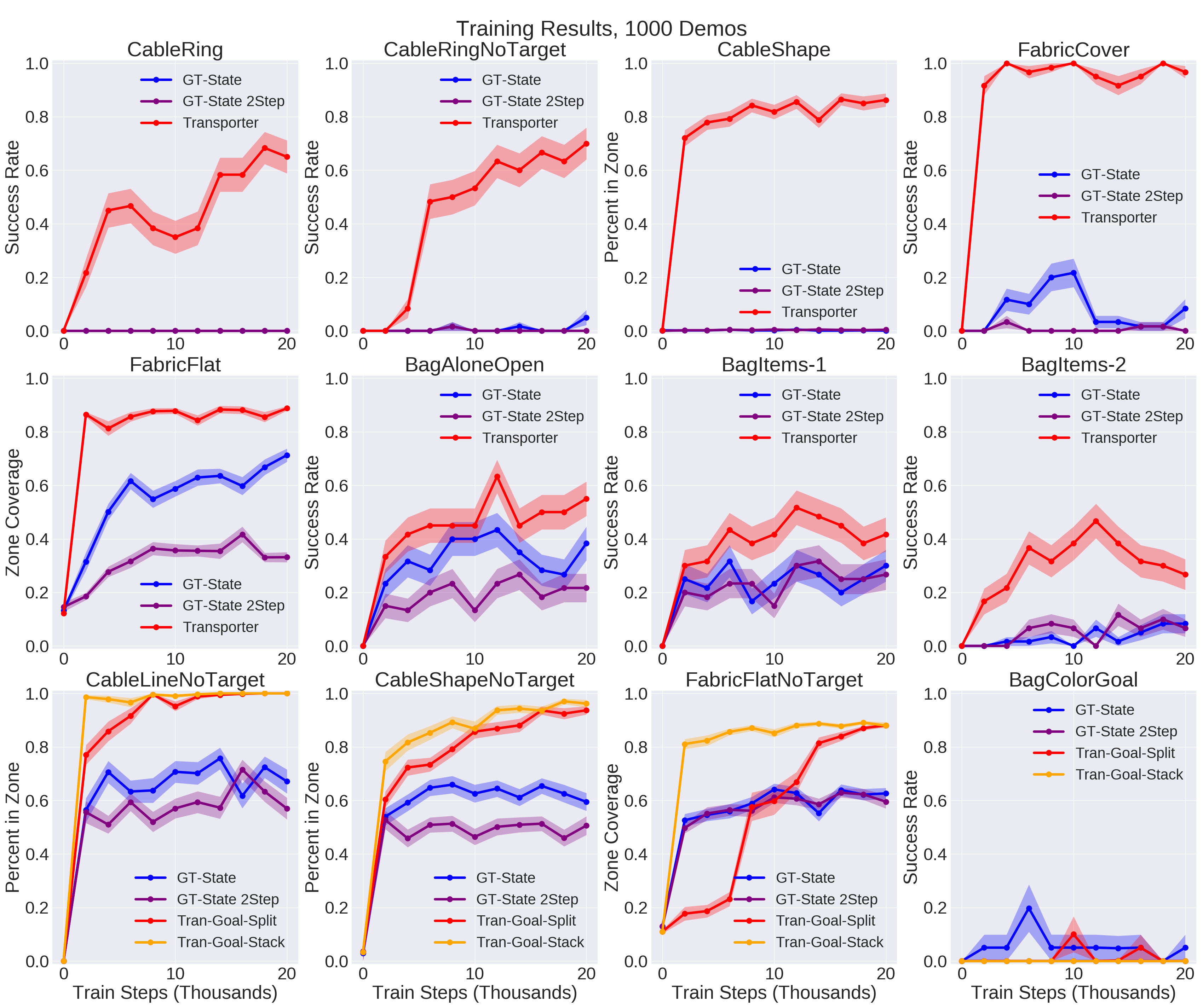}
\caption{
Results for various models, each trained on \textbf{1000 demonstrations}. The results and plots are formatted following Figure~\ref{fig:results-0001-demos}.
}
\label{fig:results-1000-demos}
\end{figure*}

In Figures~\ref{fig:results-0001-demos},~\ref{fig:results-0010-demos},~\ref{fig:results-0100-demos}, and~\ref{fig:results-1000-demos}, we show similar learning curves for the tasks in \emph{DeformableRavens} (see Table~\ref{tab:tasks}), where model architectures in each figure are all trained on the same set of 1, 10, 100, and 1000 demonstrations, respectively. 
\hl{We keep experimental settings consistent across Figures}~\ref{fig:results-0001-demos},~\ref{fig:results-0010-demos},~\ref{fig:results-0100-demos}, and~\ref{fig:results-1000-demos}.

\end{document}